**Quantifying Scripts: Defining metrics of characters for quantitative and descriptive analysis**

Vinodh Rajan

University of St Andrews

vrs3@st-andrews.ac.uk

**Abstract**

Analysis of scripts plays an important role in paleography and in quantitative linguistics. Especially in the field of digital paleography quantitative features are much needed to differentiate glyphs. We describe an elaborate set of metrics that quantify qualitative information contained in characters and hence indirectly also quantify the scribal features. We broadly divide the metrics into several categories and describe each individual metric with its underlying qualitative significance. The metrics are largely derived from the related area of gesture design and recognition. We also propose several novel metrics. The proposed metrics are soundly grounded on the principles of handwriting production and handwriting analysis. These computed metrics could serve as descriptors for scripts and also be used for comparing and analyzing scripts. We illustrate some quantitative analysis based on the proposed metrics by applying it to the paleographic evolution of the medieval Tamil script from Brahmi. We also outline future work.

**Quantifying Scripts: Defining metrics of characters for quantitative and descriptive analysis**
Vinodh Rajan
University of St Andrews
vrs3@st-andrews.ac.uk

**Introduction**

Scripts are usually seen as simple carriers of languages. Research on scripts until recently has been minimal and niche, except for the field of paleography. Scripts are however an important part of the cultural heritage of humanity and their analysis and study requires more research. Fortunately, there is a growing interest in analysis of scripts. Altmann *et al.* (2008) explore various properties of writing systems and scripts such as complexity, ornamentality and distinctivity. Changizi *et al.* (2006) discuss the various contour configurations of written symbols and their similarity to the environment in which they were produced. They also study the distribution of the configurations of various scripts. Changizi *et al.* (2005) further discuss the character complexity and the redundancy of stroke combinations of various writing systems in human history. Traditionally, analysis and study in paleography have been mostly qualitative and also done manually. Digital paleographic methods are at present making more inroads into the field. However, applying quantitative analysis on paleographic data is not yet popular and standardized (Stokes, 2009). This is partially due to the difficulty of quantifying paleographical features, and partially due to the lack of defined metrics with theoretical and qualitative underpinnings.

Scripts, being visual representation of languages, carry both linguistic and supralinguistic information. Linguistic information is closely tied to the language(s) that the scripts represent. Properties such as the phonetic valency of a character and the grapheme-to-phoneme ratio of a script can be classified as linguistic information. As supralinguistic information, apart from the visual appearance of a character, scripts importantly encapsulate handwriting behavior. In case of a particular glyphic set, they can be considered to encapsulate the handwriting behavior of a particular scribe.

In this paper, we do not consider the phonetic information contained in scripts. The phonetic information is very tightly tied to a language and quite variable. We are more focused on scripts as a set of handwritten visual symbols. In digital paleography and manuscript studies, the visual properties of a character are often more important than its phonetic properties. In these fields it is far more important to study the handwriting features for dating characters or even assigning characters to a particular scribe. We attempt to quantify and extract such information contained within a character.

The handwriting behavior contained in scripts is complex. The inter-relation between the various properties in scripts could be studied in detail. Such analysis of inter-relations could help us to identify some salient features that define handwriting behavior in humans such as those that associate glyph production and visual appearance and also throw more light on to the production of human handwriting. Multivariate analysis on the features can also help us to identify more generic features that capture salient features of scripts. In case of paleographic scripts, especially, evaluating changes in properties over a period of time would show the general trend of convergence or divergence of features for a script and also their correlations. This is useful for human computer interaction, which requires designing gesture sets with optimal features – both visual and production.

## 2. Definitions

Let us define *script* as a cohesive set of visual characters. A *character* in turn is any written symbol. A *glyph* is a particular visual representation of a character, which may deviate considerably from the normalized form of the character. *Trajectory* is the dynamic information corresponding to pen movements of the character. A *stroke* is the primitive of the handwriting process and *composite-stroke* is that which is composed of multiple primitive strokes. A *pen-drag* is movement of the pen between the intermediate pen-up and pen-down events in multistroke characters.

Let the term *metric* denote the measure that attempts to quantify a particular property of a character. We can divide metrics into two types – *absolute* & *derived*. Absolute metrics are derived directly from the structure of a character based on a particular property or feature. In many cases such as those involving length these are not scale invariant. Therefore it may be required to normalize them before some statistical operations. Derived metrics are often ratios between two absolute metrics. This is very similar to dimensionless numbers, which play an important role in several fields of engineering. The basic premise is that ratios of two metrics capture information that is more helpful than the individual metrics.

## 3. Representation of Characters

Before we turn to various metrics and their derivations, we discuss the representation of characters from which the metrics are derived.

Computationally, the "static" shape of a character is represented as a set of B-splines. B-splines are mathematical objects often used to represent complicated curves. They are known to accurately represent the curvature and shape of handwritten characters (Morasso *et al.*, 1982). Additionally, they can be manipulated with minimal effort and it is computationally easy to derive properties from B-spline representations. This conversion of glyphic shape of a character to B-splines can be done automatically or manually. In a manual process, the user defines each shape of a character directly using a set of B-splines, or explicitly draws the shape (for instance, using a drawing tablet), which is then internally converted into B-splines. An automatic conversion of a character involves thinning and then its conversion into splines. Overall, the entire process results in representation of the character's shape as a set of glyphic segments through B-splines. Figure 1 shows a character represented as a set of 5 glyphic B-spline segments.

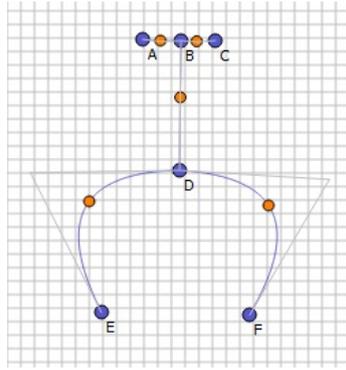

Fig. 1 Representation of a character as a set of 5 B-splines

We consider the handwritten production of characters to be the fundamental information that is contained within that character. This requires that the character be decomposed into its primitives namely strokes, which are involved in the production of characters. This is consistent with the way that the character is internalized and produced by humans. Edelman *et al.* (1987) decompose characters into four different basic template strokes - *hook, cup, gamma* and *oval*. We feel such pre-defined decomposition do not result in the creation of proper primitives. Writing is a natural process consisting of individual unique strokes, which cannot be reduced to a set of predefined templates. Changizi *et al.* (2006) decomposes the characters into "separable strokes" using three *subjects* who decide (unanimously) on the decomposition. An objective decomposition of the characters would be much more theoretically valid and also reproducible rather than relying on some underlying unknown subjective criterion.

We perform such an objective decomposition of characters into their primitive strokes using the written trajectory of the characters. For contemporary scripts the trajectory information is often known. But with paleographic scripts the trajectory is usually unknown but can be reasonably reconstructed with computational methods using their static shape (Doermann, 1992). By conducting a global search with a set of heuristics that attempts to minimize the effort required to produce the script, we attempt to reconstruct the trajectory (Jäger, 1996). Especially in case of paleographic scripts, the algorithm is able to provide several alternative viable written trajectories, among which the most viable trajectory is chosen.

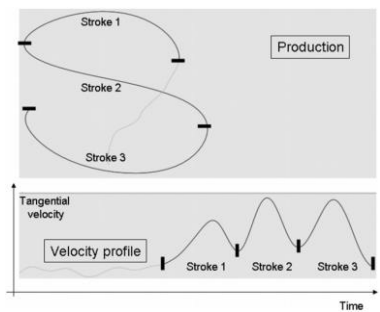

Fig. 2 Velocity profile of S and the associated stroke delineation at points of extreme curvature (Kandel et al., 2010)

Writing a character is not a discrete but rather continuous process where the individual strokes overlap and compose to form the character (Morasso *et al.*, 1981). Based on the character's trajectory, we proceed to find specific points where the (apparent) primitive individual strokes connect. Physically, handwriting is a ballistic activity with each stroke corresponding to a bell-shaped velocity profile. It consists of an acceleration phase, velocity phase and a deceleration phase (Teulings, 1993). Therefore the process of writing a character consists of several contiguous bell shaped velocity profiles corresponding to each stroke. This velocity profile of the character can be roughly predicted from the shape of the character. It is shown that the minimal velocity points occur at points where curvature is maximum or minimum (Li *et al.*, 1998) and also where strokes are explicitly delineated such as sharp junctions. The extreme points of curvature are automatically detected, and if necessary can also be manually overridden. The character is then segmented to basic strokes at all these points where the strokes overlap and/or connect, which we refer to as *Landmark Points*. In this way, we produce a natural set of primitive strokes unique to each script. This also results in the creation of a stroke-inventory for that particular script, which can be used for other types of analysis. In case of multi-stroke characters, the pen-drag between the individual strokes is included as an additional invisible stroke as it involves movement of hands also. Figure 2 illustrates the segmentation of 'S'.

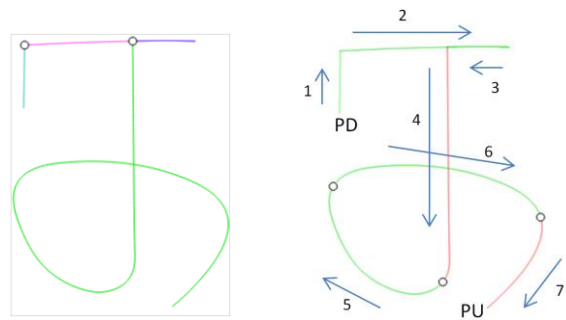

Fig 3 and 4. Character with bounding box along with the points where the strokes are disjoined and Character with the written trajectory of the character along with points where primitive strokes

This results in a new but more natural stroke-based representation of characters, which are aptly derived from their trajectories. In this new representation, the characters are composed of strokes rather than glyphic segments. The stroke primitives are also represented as B-splines similar to the glyphic segments in the original input shape. It is from this new stroke based representation that the metrics are computationally derived. Figures 3 and 4 illustrate decomposition of a character.

It is also quite possible to derive features based on a pixelated or even a contour representation of a character but we consider that not to be an accurate and natural representation of a character. Features derived from these representations such as pixel density do not accurately represent the process behind the production of the character. They are more focused towards the visual aspect of the character and the features derived are more subtle and nuanced, which may be more suitable to machine recognition. Such metrics cannot always be correlated with some explicit qualitative features as perceived by humans. The stroke-based representation of the character is much more natural and apt representation of the character, capturing accurately both the static and dynamic information contained within the character.

## 4. Information in Characters

As discussed earlier, characters contain several kinds of information within them pertaining to their production, appearance and cognition. We attempt to extract this information from a character.

Previously, attempts have been made to extract "features" in the context of pattern recognition. Rubine (1991) proposes a set of 14 "features" for the purpose of pattern recognition in gesture recognition systems. Long *et al.* (2000) similarly define 22 "features" expanding on the original features proposed by Dean (1991). Willems *et al.* (2008) perform a very elaborate literature review on different types of features and propose several new features of their own. They elaborate on 90 different features in total for online unistroke pen-gesture recognition. This entire feature set was later distilled to 49-base features that are optimal for online symbol recognition (Delaye *et al.*, 2013). Willems *et al.* (2009) also suggest additional features that pertain to multi-stroke pattern recognition. It can be seen though there is plethora of features but none of them is particularly aimed at quantifying any specific property of characters and provides substantial elaborate qualitative underpinnings for those features in terms of handwriting. They were mostly proposed as pure "statistical" features to construct feature vectors for pattern recognitions systems. However, there were some preliminary applications of these features for semantic analysis. Long *et al.* (2000) use their proposed features to analyze the subjective similarity of the gestures but they do it indirectly using multi-dimensional scaling (MDS), while Vatavu *et al.* (2011) used some of the features to correlate with "perceived execution difficulty" of gestures.

We carried out a survey of all such features used in the field of handwriting analysis and gesture analysis as listed above. We carefully studied each of the features and chose the metrics that have qualitative real-world significance and that can be directly related to properties of characters. We rejected very abstract mathematical features that do not have any direct qualitative significance. Many of the rejected features were artificially created as part of "feature construction" to increase the number of features in the feature vector, which is a common process in the field of pattern recognition. This process of "feature construction" is usually performed by applying mathematical functions such as logarithms on some basic features.

*4.1 Visual Information*

Visual information directly pertains to the appearance of a character. The following metrics are solely derived from the static shape of the character. The metrics defined below also partially quantify some production properties along with the visual property of the character.

*4.1.1 Length*

This is the total length of the character. In case of unistroke characters, this is calculated as the sum of the individual stroke lengths. For multistroke characters, this includes the previous sum and also the movement during the pen-drag, which is approximated to a straight line. *Length* quantifies the entire movement of the pen required to produce the character.

*4.1.2 Divergence*

*Divergence* is defined as distance between the position of the first pen-down event and the last pen-up event. This metric quantifies the movement of the pen between those two events measuring how much the pen has visually "diverged" from its original starting position. This is one of the important features that could be specific to a scribe.

*4.1.3 Size*

Size is measured by the bounding box area of a character. The bounding box is the minimal rectangle that encloses the given character (See figure 3). This could be directly correlated with the "largeness" of the character and hence the term "size". The bigger the bounding box, the larger is the size of the character.

*4.1.4 Length-Breadth Index*

This is the ratio of the bounding box's height to the bounding box's width. This approximates the shape aspect of the character i.e. slender/broad etc.

*4.1.5 Average Curvature*

This metric is calculated by averaging the curvature at all points of the character's strokes. The B-spline representation allows easy calculation of the curvature at each point of the curve. A straight stroke will have a curvature of zero compared to a curved stroke, which will have a higher curvature. Thus curved characters tend to have a higher average curvature compared to a character with less curves and/or more straight lines.

*4.1.6 Compactness*

This is a derived metric. Compactness of a character is defined as the ratio between *length* and the *size*. In some sense, it defines how compact (or dense) a character appears and directly corresponds to the number of strokes that a scribe is trying to fit within a given area. This makes it a very interesting metric to consider with characters. Some scribes may space out the character during production while others may tend to "compact" the strokes within a small area.

*4.1.7 Openness*

This is also a derived metric. Openness of a character can be defined as the ratio between *divergence* and *length*. This measures the movement of the pen with respect to its starting point and ending point and the length of the character. We could study the ending point of the character being varied with the length of the character. The actual metric suggested by Long *et al.* (2001) is the ratio of divergence to the size. However, this does not appear to be very ideal. It is better to compare different aspects of pen-movements (rather than the area).

*4.1.8 Distinctivity*

Several ways have been proposed to compare the appearance of characters. Jan Macutek *et al.* (2008) propose a very idiosyncratic way of calculating the distinctivity of the characters, which involves decomposing characters into basic templates and then comparing the permutations of the decomposed components. In several OCR techniques, pixel based techniques such as the image distortion model are frequently employed to calculate the similarity between the characters. Similarity (or lack thereof) is usually calculated using the cost of transformation between two entities. Entities possessing similar representations are readily transformed into one another, whereas transforming between dissimilar entities requires many transformations (Hahn *et al.*, 2003). Thus the distinctivity between characters is directly proportional to the transformations required to make them similar.

We propose to use the Dynamic Time Warping (DTW) distance (Muller, 2007) to calculate the distinctivity between two characters. DTW is traditionally employed to compare two temporal sequences, which may vary in time or speed. This makes an ideal metric to measure the difference between two characters. DTW attempts to align two sequences and calculates the cost for the alignment. The higher the DTW cost the more distinct are the signs from each other. This measure can be applied on the trajectory data or the static data. The former gives the distance based on appearance, the latter on production.

*4.1.9 Ascendancy & Descendance*

Some scripts have baselines. The percentages of the length of characters above and below baselines are defined as *Ascendancy* and *Descendance* respectively.

*4.1.10 Circularity & Rectangularity*

In many cases, the shapes of the characters appear to approach an ideal geometric shape. We attempt to measure such approximations. Circularity and Rectangularity could be defined as the deviation of the character's outline shape from that of an ideal circle and rectangle respectively. For circularity, we take the ratio of the area of the convex hull and the area of the minimal circle that encloses the character. Similarly, rectangularity can be calculated from the ratio between the area of convex hull and that of the bounding box.

*4.2 Visual Complexity*

Visual complexity can be defined as the effort required to decode and to recognize a given sign (Kohler, 2008). Some characters are perceived as complex and others as simple. Altman (2004) has proposed a technique, in which a character is decomposed into lines, arches and curves with each component assigned a weight. The sum of the weight is calculated as the quantified complexity. Peust (2006) has also proposed a complexity measure by counting the number of intersections that a character has with a straight line. These techniques do not appear to be rigorous however and are not supported by any empirical studies. Similarly, using structural information theory (SIT) there have been proposals to quantify the "load" of a character. The higher the "load", the more complex the character is to be perceived. It involves measuring repeating patterns and weights being assigned to angles of junctions (Hanssen *et al.*, 1993). While SIT can easily work for simple geometrical shapes, extending them for complex shapes such as characters is very hard and possibly not very practical.

The methods described previously attempt to quantify a very abstract notion, namely the "Visual Complexity". It is a very subjective measure as compared to others. People with exposure to different writing systems could quantify the complexity of a character in very different ways. Hence instead of aiming for complete quantification of character complexity, we propose to quantify only the factors that contribute to the visual appearance of a character. Using multidimensional techniques such as parallel co-ordinates we could trace the change in factors that contribute to the visual appearance. Along with the previously listed factors, we list also the following factors, *sum of inter-stroke angles* and *number of crossings*, which may contribute to visual complexity.

## 5. Dynamic Information

Apart of the static shape of a character we need to also consider its dynamics. The character's kinematic (or temporal) information is essential in defining it. It dictates how the character is produced through the process of handwriting. Thus, deriving metrics quantifying properties of its production is very much important.

*5.1 Stroke Counts*

It is a fundamental metric to count the number of hand motions required to write the characters. Humans consistently attempt to minimize the number of hand-movements to write characters (Saloman, 2012). It is an interesting metric to analyze for the distribution across various scripts. Apart from the count of the primitive strokes, there are two more composite-stroke metrics that could be considered – pen-strokes & disjoined strokes. The former is the absolute hand movements required to write characters without a pen-up even and the latter is the composite-strokes that are delineated at sharp-junctions. For instance, figures 3 & 4 show a character with 1 pen-stroke, 3 disjoined strokes and 8 primitive strokes. We could also include *retraces* in the count, where the same stroke is traced successively in the opposite direction. Movement *3* in figure 4 is the retracing stroke.

*5.2 Stroke Length*

The distribution of the length of individual strokes and also calculating the average stroke length is a very purposeful measure with respect to the analysis of writing. The average stroke length is a variable entity across different scripts or scribes.

*5.3 Changeability*

Handwriting consists of up-strokes and down-strokes. They are of two different characteristics with completely different physiological process of production. It has been shown up-strokes are susceptible to change, while down-strokes are considered invariant (Teulings, 1993) and more stable (Maarse 1983). Upstrokes are faster (Isokoski, 2001) and hence perhaps less stable. Maarse *et al.* (1983) defines strokes that are produced between 210° and 280° to be downstrokes. The range of angles appears to be very restrictive (as it considered only Roman handwriting). Hence, we have included strokes which are pointed downwards within 210° and 330° as down-strokes and all non-down strokes are included as up-strokes. So the ability of the character to change i.e. *changeability* can be directly tied to the ratio of upstrokes' length to that of the down-strokes' length. If the ratio is high the character can be considered susceptible to change. Thus changeability as a metric is related to a character's susceptibility to change.

*5.4 Disfluency*

As discussed earlier, writing is a ballistic activity. It is known that handwriting fluency is affected at points where curvature is at its maximum/minimum. The transition between down-strokes and up-strokes is also considered to slow down the writing process. The number of sharp junctions in a character also contributes to the slowing of velocity during the handwriting production. The sum of all points that affect velocity – *curvature extrema, sharp-junctions, and intermediate pen-up events* - is termed as *disfluency*. This can directly correspond to the difficulty in terms of writing the character. A character with higher number of disfluent points is harder to produce as the velocity is frequently interrupted. Similar measures have been used with actual dynamic handwriting velocity data to assess handwriting fluency of people by measuring the number velocity slow-downs happening (Tucha, 2008). Character in figures 3 & 4 has 6 disfluent points.

In fact, the number of disjoined points can be taken as separate metric all together, since its effect on slowing down the production is higher than that of the other points.

*5.5 Entropy*

In information theory, entropy is defined as the average amount of information contained within an entity. This amount of information in the system is directly proportional to the randomness or disorderliness present in the system. When there are several instances of change, it results in increase of entropy as it contains more information (Aksentijevic *et al.*, 2012). To calculate the entropy of a character the trajectory of the character is "quantized" into Hoffman codes denoting the major eight directions. Assigning a Hoffman code to the individual strokes performs this. The eight Hoffman codes correspond to the following directions - N, S, E, W, NE, NW, SE, and SW. The sample character in Figure 3 can be quantized into [N E W S NW SE SW].

Entropy is calculated based on the following formula (Bhat *et al.*, 2009):

$H(s) = \Sigma\ p(s_i)\ \log_e\ p(s_i)$

Where, $p(s_i)$ is the probability of a stroke. It is given by ratio of the count of the given stroke (in the character) to that of the total number of strokes.

Any character with a sufficient number of repeating patterns will record low entropy and those with no patterns high entropy. Thus the entropy of characters conveys the randomness associated with the pen movements required to produce the character.

*5.6 N-Gram model of scripts*

Writing a character can be very well considered to be similar to that of constructing a sentence. While sentences are made up of words, characters are made of strokes. We here seek to apply some aspects of natural language processing to scripts. N-gram modeling is frequently used in natural language processing for a wide variety of purposes. N-gram model is a probabilistic model to predict the next item in a sequence (Fink, 2014). As the number of stroke combination is usually low, a bigram model would better to model script behavior. The n-gram model provides an opportunity to derive several metrics. It is now possible to calculate the entropy of a script as opposed to that of a character and also allows us to study the regularity of stroke combinations.

*5.7 Angle-Based Metrics*

Analyzing the different angles of strokes occurring in scripts can throw more light on a particular scribal behavior. We define a few important angle-based metrics that could be used. *Major Angle* would be the angle of the major primitive stroke present in the character. The *initial angle* is defined with the initial stroke of a character. The *divergence angle* - Angle between first and last points could also be considered as a metric. For multi-stroke characters, *angle of pen-drag* can be an important measure. *Inter-stroke angles* could be plotted as a histogram to see the changes.

*5.8 Pen-Drag Distance*

The *Pen-drag Distance* is a metric with respect to multi-stroke behavior. This captures the hand movements between pen-strokes, which are an important part of multi-stroke production.

**6. Cognitive Information**

Writing a character is usually a top-down process. A character has to be memorized and then reproduced. Consequently, this requires elaborate trajectory planning. We need to find out the approximate information required to cognitively memorize and produce characters.

In this respect, we refer to the Algorithmic Information Theory (AIT). Especially within AIT, Kolmogorov complexity attempts to find the minimal description of a given sequence (Wallace *et al.*, 1999). In a similar way, we attempt to find out the minimal representation of a character required to reproduce it. Theoretically, these would be the points necessary to plan the trajectory of the character. In fact, this directly corresponds to the "Landmark Points" in the character, as those are the points that define its shape. Isokoski (2001) measured the complexity of characters, by studying the number straight of lines required to approximate a character. But again this was a subjective measure. In lieu of this, the number of landmark points required in a character could be considered a direct theoretical implementation of Isokoski's complexity. Do note that this is an approximation. The proximity and distribution of the landmark points may affect the information contained in the character (for instance, if several points are very close to each other it might create additional confounding factors to the planning of the trajectory, which will increase the information content). But for now, we ignore such intricate details. These need to be studied in detail later.

The Ramer-Douglas-Peucker (RDP) algorithm (Douglas *et al.*, 1956) also computes the *minimum number of points* required to approximate a given curve. This mostly agrees with the number of landmark points in some case, but in other cases this might not be case so. The issue with RDP is that a threshold for approximation needs to be provided and it might provide a slight over-estimation of the points required for approximation.

Both the *RDP* and *Number of Landmark points* can be considered as different metrics that correspond to the cognitive information present in a character and could be used as required.

**7. Metrics of Scripts Vs. Metrics of Characters**

Most of the metrics discussed in the paper were confined to that of individual characters. However, as defined earlier, a script is a cohesive set of characters. In many cases, the metric of the script could be found just by averaging the metric of the individual characters. For instance, it would be possible to discuss the average curvature of a script and even compare them. Characters within a script are usually a heterogeneous set with different purposes and different patterns of usage. Hence, an average metric for the script may not always make sense. In such cases, instead of averaging the metrics, it is more useful to study the distribution of a metric in different scripts. Since characters within a script behave as a set, studying the homogenization (or divergence) of properties within the script is a useful exercise. It would also be more useful if this could be overlaid with some other information such as usage frequency of the characters. For instance, an interesting analysis would be to see how the various properties of frequently used characters differ with respect to rarely used characters or indeed, if any such difference exists at all.

**8. Development of Medieval Tamil from Brahmi – A Quantitative Analysis**

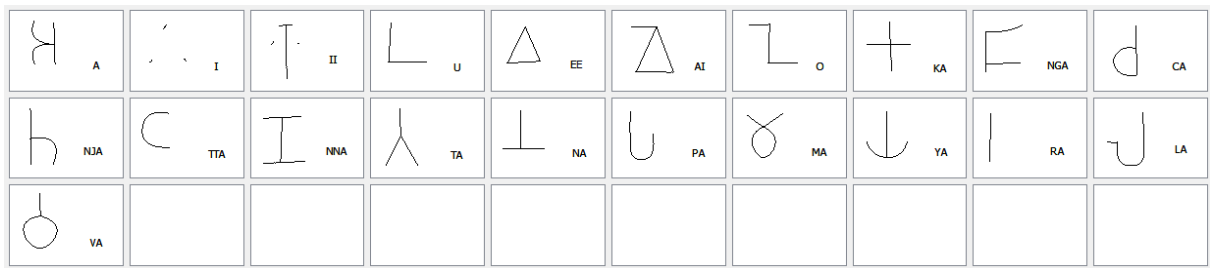

Fig. 5a Digitized Tamil 1 Script

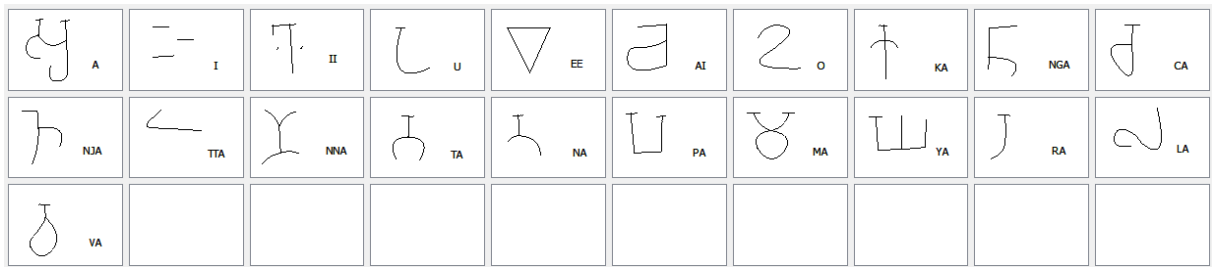

Fig. 5b Digitized Tamil 2 Script

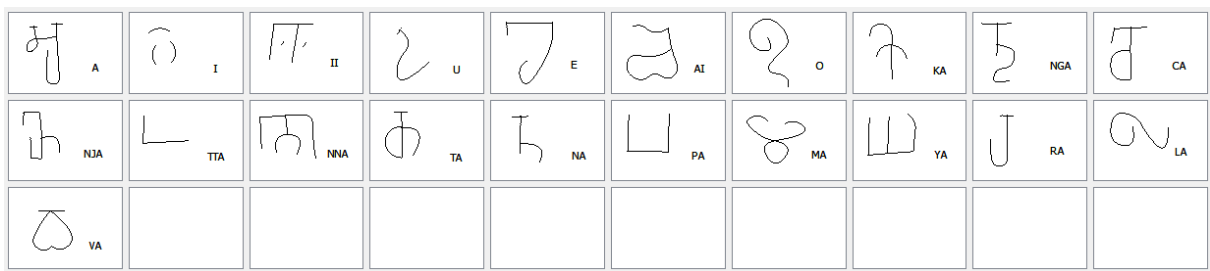

Fig. 5c Digitized Tamil 3 Script

For illustrating the quantitative analysis that can be performed with the proposed metrics, we take the medieval Tamil script in 3 different stages of evolution beginning from Brahmi (Ojha, 1964). We have developed a prototype application that implements the discussed representation of a character and derives appropriate metrics from characters. The paleographic characters were digitized using our framework and stored. We then proceeded to reconstruct the trajectories, to segment them into strokes and to extract the required metrics. We have attempted to use some of the major metrics discussed to keep the section succinct. We hope that paleographers and quantitative linguists may find various other ways to use these metrics (and other proposed metrics) to support their analyses as they may require. Figures 5a, 5b and 5c show the medieval Tamil script in 3 different stages of evolution. We refer to them as Tamil 1 (figure 5a), Tamil 2 (figure 5b) and Tamil 3 (figure 5c) respectively.

*8.1* Changes in Visual Features

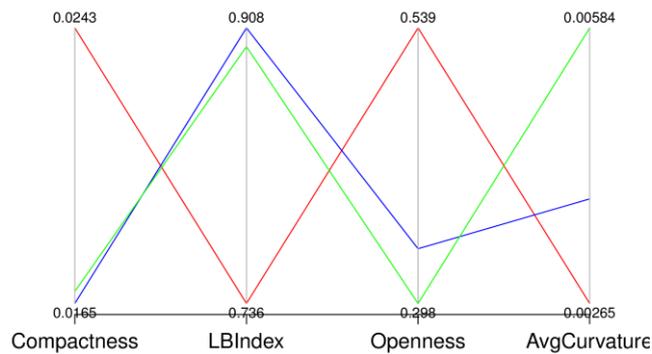

Fig. 6 Parallel co-ordinate plot for mean values of the different visual features – Tamil 1 (Red), Tamil2 (Blue) and Tamil 3 (Green)

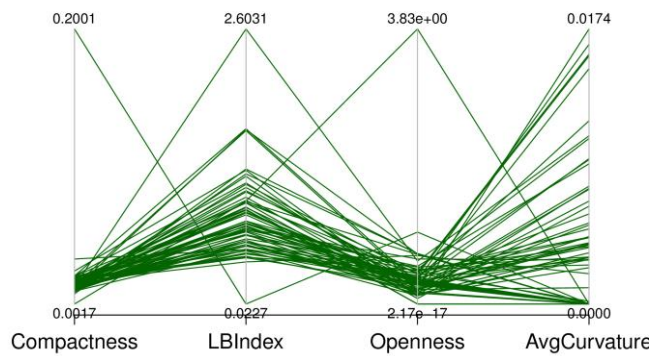

Fig. 7: Parallel Co-ordinate plot for the entire character set in Tamil 1, Tamil 2 and Tamil 3

Figure 6 shows the parallel co-ordinate plot for mean values of the different visual features of the 3 scripts. Tamil 1 starts out with the base arrangement of high compactness and openness and a low curvature with the characters being more slender (as noted by the low LB Index). But inadvertently the features appear to have quite divorced from their initial appearance. They have become more and more symmetrical sized, less compact, with strokes being augmented and less open with increase in average curvature. One interesting point here to be noted is that Tamil 2 starting leaning towards being curved and ended up becoming extremely curved by a very large magnitude as much as Tamil 3. This probably resulted from the scribes trying to make the script look more elegant or perhaps due to the change in implement (and writing material). From figure 7 it can also be seen that compactness and openness appear to be tightly related (at least for the scripts under discussion).

*8.2 Distribution of Entropy*

Figures 8a, 8b and 8c show the distribution of entropy across the scripts as histograms. As the script evolved over time, it appears to have gained information through stroke augmentations, hence the change in distribution skewing towards the right (See Tamil 2). But in the final medieval version of the script, some information content gained appears to have been lost. This can be attributed to the fact that the stroke patterns were later developed. Also, in some characters some strokes were later lost.

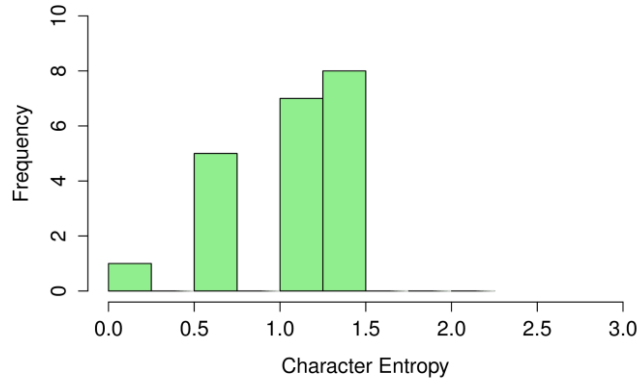

Fig 8a: Distribution of Character Entropies for Tamil 1

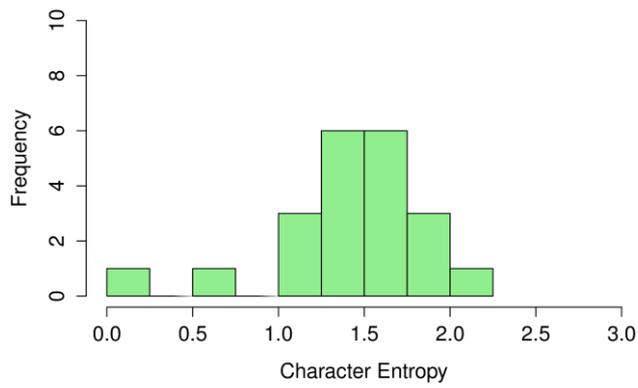

Fig 8b: Distribution of Character Entropies for Tamil 2

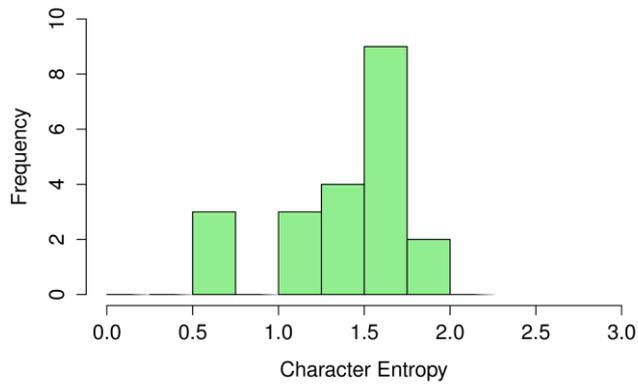

Fig 8c: Distribution of Character Entropies for Tamil 3

*8.3* Stroke Count

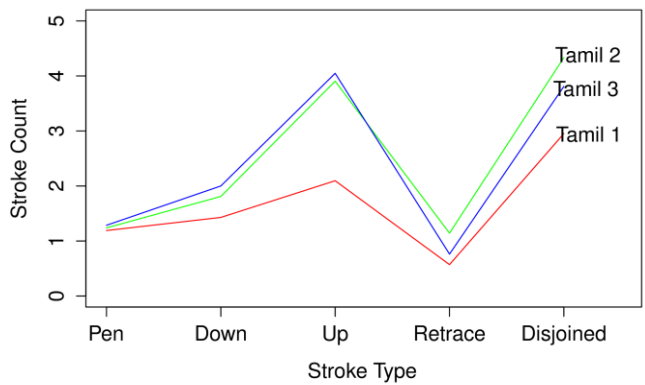

Fig 9. Line plot comparing the stroke counts

Line plot comparing the different stroke counts is shown in figure 9. The number of pen-strokes has remained nearly constant. There appears to have been a surge in the up-strokes but not that of that down strokes. It is consistent with the fact that the downstrokes are more stable (probably are more hard to produce) and hence do not show an increase as compared to the upstrokes, which are more fluid.

*8.4 Distinctivity*

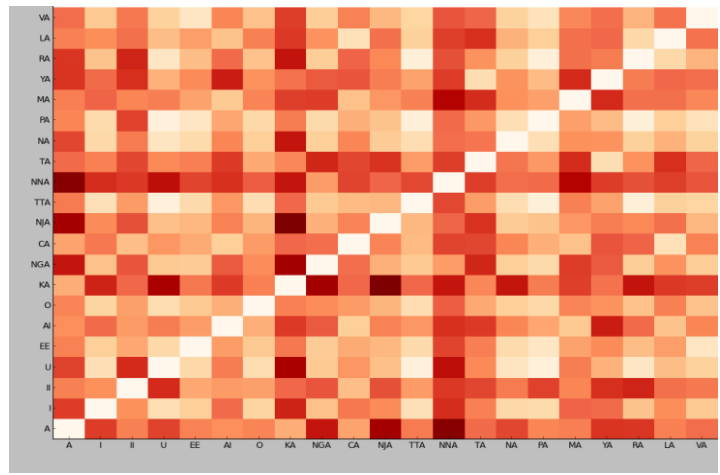

Fig 10a. Heatmap of Distinctivity for Tamil 1

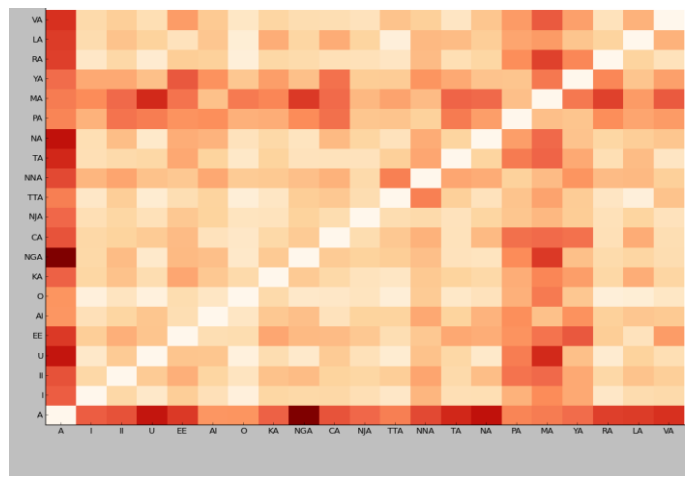

Fig 10b. Heatmap of Distinctivity for Tamil 2

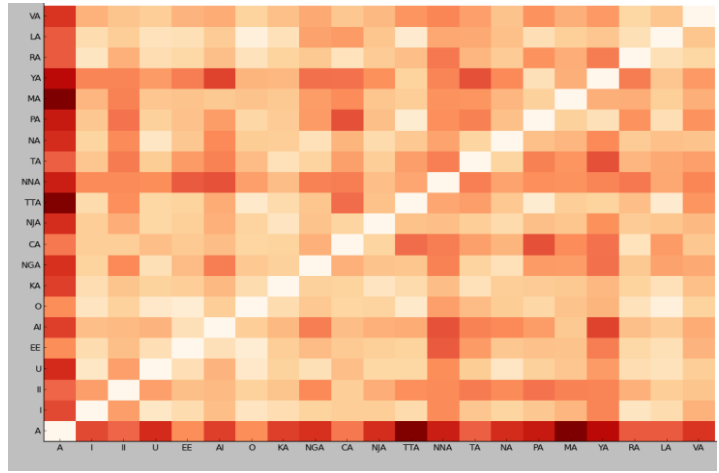

Fig 10c. Heatmap of Distinctivity for Tamil 3

We attempt to study the change in distinctivity of scripts. We first construct a self-similarity matrix using the computed distinctivity values. Using the self-similarity matrix, we then generate a heatmap for the script. Figures 10a, 10b and 10c show the generated heatmap for Tamil 1, Tamil 2 and Tamil 3 respectively. The lighter shades show the least distinctivity while the darker shades more distinctivity. It can be seen that the base version of the script (Tamil 1) started out with the characters being very distinct from each other. Later, the characters proceed to become very self-similar, only to get gain more distinctivity later. This throws a bit of light on the evolution of scripts (at least of Tamil in this case).

It can be inferred from the pattern that, if characters within a script get more and more similar to each other they contribute more to the confounding factor. To counteract the increasing confounding nature of scripts, their characters tend towards more disambiguation. This appears to be a valid inference for the behavior of scripts.

*8.4 Major Pen Direction*

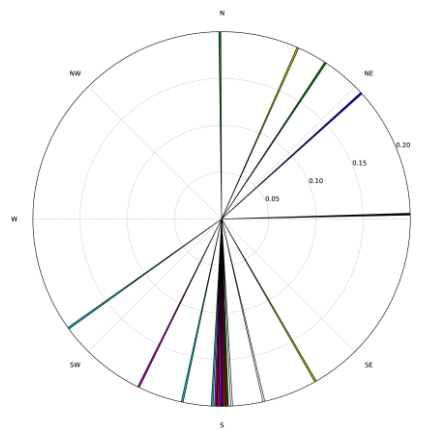

Figure 11a: Visualization of Initial pen direction for Tamil 1

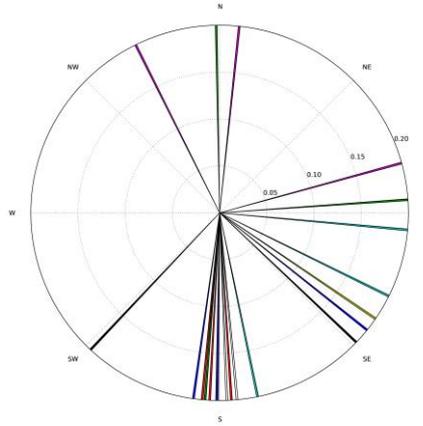

Figure 11b: Visualization of Initial pen direction for Tamil 2

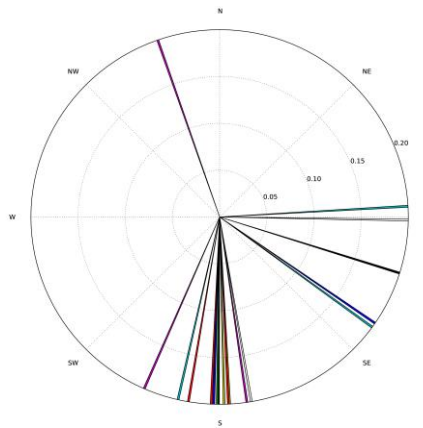

Figure 11c: Visualization of Initial pen direction for Tamil 3

The directions of the major strokes in the different scripts are shown in figures 11a, 11b and 11c for the 3 scripts. In terms of the major pen Direction, it can be seen that has always stayed consistent towards the downward direction during the evolution of the medieval Tamil script.

## 9. Future Work

The metrics were aimed to be used for descriptive and exploratory analysis. But one of the most important applications of these metrics would be in digital paleography to classify glyphs with some sound quantitative support. These metrics could provide both quantitative and qualitative reasoning behind such classification. As a part of future work, we intend to evaluate the effectiveness of the proposed metrics in classification of characters such as assigning the character to a particular scribe, time period or even a script. This would throw some light on the effectiveness of these metrics in classifying hands. Even if not for complete automatic classification, the amount of assistance these metrics could provide paleographer sand manuscript experts with such classification is also an interesting problem to investigate. We intend to perform such evaluation in the future.

## 10. Conclusion

We have discussed the necessity of well-defined metrics for characters. We then defined a framework for a natural representation of characters using the primitive "strokes" to enable extraction of metrics. The various kinds of information contained in the characters have been elaborated. We then proceeded to define and elaborate on various metrics under thee different categories – visual, production and cognitive. To illustrate the metrics we attempted to perform quantitative analysis on development of Medieval Tamil script for Brahmi using some of the proposed metrics. We have quantitatively described the process of script evolution and have derived some interesting inferences from the analysis. As a part of the future work, we discuss the possibility of empirically evaluating the effectiveness of the metrics for classifying characters as opposed to the human classification.